\documentclass[a4paper, 10pt, conference]{ieeeconf}      

\IEEEoverridecommandlockouts

\usepackage{graphics} 
\usepackage{epsfig} 
\usepackage{amsmath} 
\usepackage{amssymb}  
\usepackage{bm}

\usepackage{blindtext}
\usepackage{hyperref}
\hypersetup{colorlinks, 
  linkcolor=red,
  citecolor=blue,
  urlcolor=red}  

\usepackage{lipsum}

\usepackage{array}
\newcolumntype{P}[1]{>{\centering\arraybackslash}p{#1}}
\newcolumntype{L}[1]{>{\arraybackslash}p{#1}}
\usepackage{multirow}

\usepackage{subcaption}
\usepackage{siunitx}
\usepackage{booktabs}
\usepackage{tabularx}

\usepackage{dirtree}

\usepackage{csquotes}
\usepackage[url=false, doi=false, isbn=false, sorting=none]{biblatex}
\usepackage[printonlyused]{acronym}
\acrodef{ICP}{Iterative Closest Point}
\acrodef{lpm}{libpointmatcher}

\usepackage{float}

\title{\LARGE \bf When and Where Localization Fails: An Analysis of the Iterative Closest Point in Evolving Environment}

\author{Abdel-Raouf Dannaoui$^{1,2}$, Johann Laconte$^{1}$, Christophe Debain$^{1}$, Francois Pomerleau$^{2}$ and Paul Checchin$^{3}$%
  \thanks{$^{1}$Université Clermont Auvergne, INRAE, UR TSCF, 63000, Clermont-Ferrand, France
    {\tt\small abdel-raouf.dannaoui@inrae.fr}}%
  \thanks{$^{2}$Northern Robotics Laboratory, Universite Laval, Canada.}%
  \thanks{$^{3}$Institut Pascal, Université Clermont Auvergne, Clermont Auvergne INP, CNRS, F-63000 Clermont-Ferrand, France}
}

\begin{document}

\maketitle
\thispagestyle{empty}
\pagestyle{empty}

\begin{abstract}%
  \label{sec:abstract}
    Robust relocalization in dynamic outdoor environments remains a key challenge for autonomous systems relying on 3D lidar. While long-term localization has been widely studied, short-term environmental changes, occurring over days or weeks, remain underexplored despite their practical significance. To address this gap, we present a high-resolution, short-term multi-temporal dataset collected weekly from February to April 2025 across natural and semi-urban settings. Each session includes high-density point cloud maps, \SI{360}{deg} panoramic images, and trajectory data. Projected lidar scans, derived from the point cloud maps and modeled with sensor-accurate occlusions, are used to evaluate alignment accuracy against the ground truth using two \ac{ICP} variants: Point-to-Point and Point-to-Plane. Results show that Point-to-Plane offers significantly more stable and accurate registration, particularly in areas with sparse features or dense vegetation. This study provides a structured dataset for evaluating short-term localization robustness, a reproducible framework for analyzing scan-to-map alignment under noise, and a comparative evaluation of \ac{ICP} performance in evolving outdoor environments. Our analysis underscores how local geometry and environmental variability affect localization success, offering insights for designing more resilient robotic systems.
\end{abstract}

\section{INTRODUCTION}%
\label{sec:introduction}
As robotic systems progress toward greater autonomy, their ability to navigate unstructured environments remains a fundamental challenge. 
Unlike structured settings, unstructured environments are inherently complex, evolving, and unpredictable. 
They feature uneven terrain, varying weather, and limited prior knowledge, all of which complicate perception, localization, and planning~\cite{bib:barilKilometerscaleAutonomousNavigation2022}. 
This challenge becomes particularly prominent in applications such as agriculture, where robots must navigate dense vegetation and uneven terrain~\cite{bib:liReviewHighThroughputField2022}. 
Similarly, in environmental monitoring, robots often operate in complex, unstructured settings such as forests and mountain regions~\cite{bib:pierzchalaMappingForestsUsing2018, bib:nieForest3DLidar2022}. 
Even in urban areas, autonomous vehicles face the need to manage dynamic obstacles and degraded GPS signals~\cite{bib:sanchezRobustNormalVector2020}.

\begin{figure}[t!] 
  \centering
  \includegraphics[width= \linewidth]{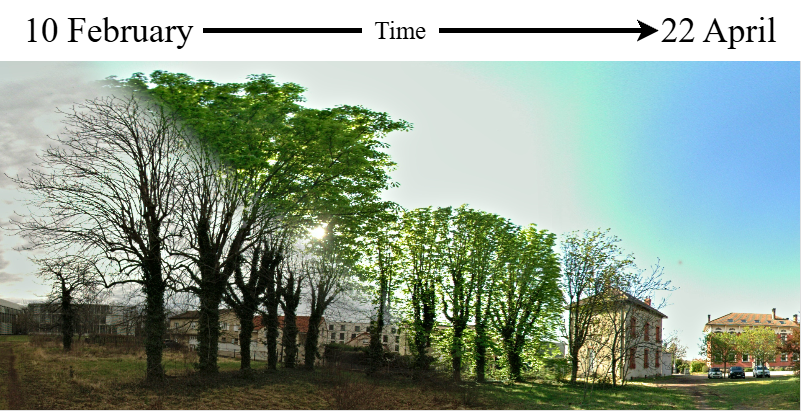}
  \caption{Visual progression of the natural environment from the beginning of the dataset to April 22. The scene illustrates the seasonal evolution of vegetation, showing trees transitioning from bare branches to full leaf coverage over time. Such changes will heavily modify the perception of the environment.}%
  \label{fig:ecmr-head}
\end{figure}

Relocalization, the process by which a robot determines its position within a known map, is crucial for a variety of tasks, such as autonomous inspection. 
Advances in Simultaneous Localization and Mapping (SLAM) and map-based relocalization have significantly improved performance, particularly for long-term autonomy and teach-and-repeat frameworks~\cite{bib:peng_roll_2022}. 
However, as illustrated in \autoref{fig:ecmr-head}, changes in the environment, such as foliage growth, seasonal transitions, and moving objects, can cause sensor-to-map mismatches, degrading localization performance. 
Specifically, questions such as \emph{Which changes most affect relocalization?}, \emph{Can we predict failure-prone areas?}, or \emph{Where do current systems remain unexpectedly robust?} are still underexplored.
Most benchmarks either cover single-day acquisitions~\cite{bib:pomerleau_comparing_2013} or wide seasonal gaps~\cite{bib:pomerleau_challenging_2012}, missing the nuance of short-term environmental evolution.

To address this gap, we introduce a high-resolution, short-term multi-temporal dataset recorded weekly from February until April 2025,  spanning both natural and semi-urban environments. 
Unlike long-term seasonal datasets, our focus is on short-horizon changes, such as branch trimming, vehicle displacement, and vegetation regrowth.
Using this dataset, we analyze a lidar-based relocalization pipeline by aligning point clouds from a previously recorded map with observations captured in the surrounding weeks.
By applying perturbations and computing pose errors, we expose typical failure modes and non-intuitive success cases across diverse conditions. Our contributions are:
\begin{itemize} 
    \item A real-world, high-resolution, short-term multi-temporal dataset designed to study localization under environmental change;
    \item A relocalization evaluation framework using perturbed lidar scans and \ac{ICP}-based registration, quantifying degradation over time; and 
    \item An error-driven analysis revealing both expected and unexpected relocalization behaviors in real environments. 
\end{itemize}

\section{RELATED WORK}%
\label{sec:related_work}
\subsection{Multi-temporal Datasets}%
\label{sec:multi-temporal_datasets}
Multi-temporal datasets play a foundational role in evaluating the robustness of localization and mapping algorithms under real-world environmental changes. 
A variety of datasets presented in \autoref{tab:datasets_summary} have been introduced over the past decade, capturing diverse conditions ranging from long-term seasonal shifts to more dynamic, short-term scene variations.

In urban and semi-urban contexts, NCLT~\cite{bib:ncarlevaris-2015a}, Boreas~\cite{bib:burnett_boreas_2023}, 4Seasons~\cite{bib:wenzel_4seasons_2025}, and the dataset by~\textcite{bib:pomerleau_challenging_2012} have set the standard for long-term evaluation. 
These datasets span months or even years, capturing a wide range of weather and lighting conditions for assessing SLAM and localization algorithms across extended deployments. 
Their consistent structure, rich sensory modalities, and challenging driving scenarios have made them cornerstones in research on visual and lidar-based localization.

In natural and forested environments, robustness to dense vegetation, terrain variability, and occlusion is vital.
FoMo~\cite{bib:boxan_fomo_2024}, FinnForest~\cite{bib:ali_finnforest_2020}, and Wild-Places~\cite{bib:knights_wild-places_2023} provide important testing grounds for these challenges. 
They include varied seasons, revisit loops, and in the case of Wild-Places, large-scale lidar place recognition in unstructured settings. 
These datasets emphasize the importance of environmental diversity and highlight the need for localization solutions that are resilient to both seasonal and structural environmental changes.

However, while these long-term datasets capture broad environmental evolution, they typically miss fine-grained, short-horizon variations, such as weekly changes in vegetation density, temporary obstacles such as parked vehicles, or early, stage growth. 
ROVER~\cite{bib:schmidt_rover_2025} and VPAIR~\cite{bib:schleiss_vpair_2022} provide valuable insights into long-range relocalization and aerial visual place recognition, respectively, but their revisit intervals and perspectives differ significantly from the ground-level, weekly revisits required for short-term analysis. 
Similarly, but for simulated datasets, VIODE~\cite{bib:minoda_viode_2021} enable controlled studies of dynamic scenes but lack real-world complexity. 
Other approaches, including CrowdDriven~\cite{bib:jafarzadeh_crowddriven_2021} and OutFin~\cite{bib:alhomayani_outfin_2021}, explore localization using crowd-sourced or multi-device modalities, offering breadth but not the repeatable, sensor-rich capture needed for detailed short-term mapping and relocalization performance analysis.

To address this gap, our proposed dataset introduces a high-resolution, weekly multi-temporal capture of two physically distinct environments, natural and semi-urban, over several months. 
Unlike existing datasets, it preserves record timing, consistent trajectories, accurate alignment, and dense semantic labeling across sessions. 
This enables focused investigation of short-term environmental changes on lidar-based localization performance, a dimension underrepresented in current benchmarks.

\subsection{Localization algorithms}%
\label{sec:localization_algorithms}
Despite significant progress in lidar-based localization, environmental dynamics remain a major obstacle, particularly in outdoor settings where vegetation growth, weather shifts, and object displacements can degrade scan-to-map alignment. 
Multiple studies have sought to improve robustness by incorporating motion segmentation, dynamic object filtering, or specialized scan matching techniques. 
PointLocalization~\cite{bib:zhu_real-time_2019} demonstrate that lidar-based systems can maintain resilience against outdated maps, despite challenges introduced by dynamic objects, narrow sensor fields of view, and degraded GPS in urban settings. 
Similarly, scan similarity metrics and drivable region segmentation~\cite{bib:hsu_3d_2019} have shown improvements in alignment in cluttered scenes. 
Additionally,~\textcite{bib:pomerleau_comparing_2013} presents an extensive comparison of \ac{ICP} implementations in static real-world data sets that cover a variety of environments. 
While their datasets include dynamic outliers, the primary focus is on comparing performance across different types of static environments, varying overlap ratios, and different levels of initial pose perturbation.
However, these methods fall short of pinpointing where and why localization fails in evolving environments. 
Their results highlight general robustness but lack a fine-grained, spatially-aware analysis of error.

Broader surveys~\cite{bib:elhousni_survey_2020} classify existing approaches into registration, feature-based, and deep learning methods, evaluating across standard benchmarks such as KITTI. 
Yet, they do not deeply examine the sensitivity of these methods to evolving scene conditions. 
Similarly, advances in fast registration~\cite{bib:tinchev_real-time_2023} and neural map representations~\cite{bib:zhong_3d_2024} offer practical gains by enabling real-time operation or filtering dynamic elements, but neither study systematically evaluates how specific types of environmental changes, such as occlusion, growth, or structure loss, affect localization reliability.

This lack of targeted analysis leaves a gap in our understanding: 
our work builds directly on these efforts by introducing a dataset and evaluation framework focused on short-term, real-world environmental change. 
By controlling the trajectory and recording the same information each week, we enable a precise understanding of the environment's evolution and localization robustness.

\begin{table*}[ht!] 
  \centering
  \caption{Summary of Datasets.}%
  \vspace{-0.2cm}
  \caption*{\textbullet{} Sensor is included \hspace{3cm} \textopenbullet{} Sensor is not included}
  \label{tab:datasets_summary}
  {\footnotesize 
    \begin{tabular}{l P{1.5cm} P{2.5cm} P{0.8cm} P{0.5cm} P{0.5cm} P{2cm} P{2cm} P{2cm}}
      \toprule
      \textbf{Name} & \textbf{Environment} & \textbf{Application} & \textbf{Camera} & \textbf{Lidar} & \textbf{IMU} & \textbf{Strengths} & \textbf{Weaknesses} & \textbf{GT} \\
      \midrule
      F. Pomerleau~\cite{bib:pomerleau_challenging_2012}                                                    
                & Diverse & Pointcloud registration 
                & \Large\textopenbullet{} & \Large\textbullet{} & \Large\textbullet{} 
                & Challenging features & No weekly changes
                & Theodolite \\
      NCLT~\cite{bib:ncarlevaris-2015a}                                                    
                & Urban & Long-Term Research 
                & \Large\textbullet{} & \Large\textbullet{} & \Large\textopenbullet{} 
                & Urban/Seasonal Variations & Big Time Differences
                & GPS \\
      Boreas~\cite{bib:burnett_boreas_2023}                                                
                & City & Long-term localization
                & \Large\textbullet{} & \Large\textbullet{} & \Large\textbullet{}
                & Multi-season/weather & Only in city
                & GNSS\\
      4Seasons~\cite{bib:wenzel_4seasons_2025}
                & Urban Driving & Long-Term Visual Localization 
                & \Large\textbullet{} & \Large\textopenbullet{} & \Large\textopenbullet{} 
                & Seasonal/weather Variations & No short-term revisit
                & GNSS/V-Odometry \\
      FoMo~\cite{bib:boxan_fomo_2024}                                                       
                & Boreal Forest & Multi-Season Navigation 
                & \Large\textbullet{} & \Large\textbullet{} & \Large\textbullet{} 
                & Multi-Sensor, Multi-Seasons & Specific To Montmorency Forest
                & GNSS \\
      FinnForest~\cite{bib:ali_finnforest_2020}   
                & Forest & Multi-Season Navigation 
                & \Large\textbullet{} & \Large\textopenbullet{} & \Large\textopenbullet{} 
                & Forest And Daylight Conditions & Camera-Based Measurements
                & GNSS \\
      Wild-Places~\cite{bib:knights_wild-places_2023}                   
                & Forest & Long-Term Navigation, Mapping
                & \Large\textbullet{} & \Large\textbullet{} & \Large\textopenbullet{} 
                & Loop Closure & Big Time Differences
                & GPS, SLAM\\
      ROVER~\cite{bib:schmidt_rover_2025} 
                & Natural & Visual SLAM, Place Recognition 
                & \Large\textbullet{} & \Large\textopenbullet{} & \Large\textbullet{} 
                & Seasonal variation & No short-term changes
                & V-Odometry \\
      VPAIR~\cite{bib:schleiss_vpair_2022}
                & Urban/Nature & Aerial Visual Place Recognition 
                & \Large\textbullet{} & \Large\textopenbullet{} & \Large\textopenbullet{} 
                & Varied environments & Only aerial view
                & GPS \\
      VIODE~\cite{bib:minoda_viode_2021}
                & Simulated Urban & VIO in Dynamic Environments 
                & \Large\textbullet{} & \Large\textopenbullet{} & \Large\textbullet{} 
                & Reproducibility & No real-world complexity
                & Simulated \\
      CrowdDriven~\cite{bib:jafarzadeh_crowddriven_2021}
                & Diverse Outdoor & Visual Localization Benchmarking 
                & \Large\textbullet{} & \Large\textopenbullet{} & \Large\textopenbullet{} 
                & Real-world failure cases & Uncontrolled conditions
                & Approximate pose \\
      OutFin~\cite{bib:alhomayani_outfin_2021}
                & Outdoor (GPS-Denied) & Fingerprint-based Localization 
                & \Large\textbullet{} & \Large\textopenbullet{} & \Large\textbullet{} 
                & Multi-modal  & Not lidar/SLAM-centric
                & WiFi/GPS ground truth \\
      Nordland~\cite{bib:sunderhauf_are_nodate} 
                & Train Rails & Place Recognition 
                & \Large\textbullet{} & \Large\textopenbullet{} & \Large\textopenbullet{} 
                & 4 Seasons, Synced & 1 Visit/Season
                & GPS \\
      UTIAS~\cite{bib:gridseth_keeping_2022}                                             
                & Natural & Visual Odometry 
                & \Large\textbullet{} & \Large\textopenbullet{} & \Large\textopenbullet{} 
                & Daily Runs & Only Camera
                & GPS\\
      Oxford Forest~\cite{bib:yan_long-term_2023}
                & Forest & {Lidar} Place Recognition 
                & \Large\textopenbullet{} & \Large\textbullet{} & \Large\textopenbullet{} 
                & Unstructured natural & Sparse temporal coverage
                & SLAM \\
      \midrule
      \multirow{3}{*}{\textbf{Ours}}                                          
                & \textbf{Natural} \textbf{Semi-Urban} 
                & \textbf{Short-term multi-temporal Localisation} 
                & \multirow{3}{*}{\Large\textbullet{}} & \multirow{3}{*}{\Large\textbullet{}} & \multirow{3}{*}{\Large\textbullet{}}
                & \textbf{Dense  maps, New map \hspace{0.2cm} every week} 
                & \textbf{No Raw Scans, No Night Environment}
                & \multirow{3}{*}{\textbf{Pose (GNSS/IMU)}}\\
      \bottomrule
    \end{tabular}}
\vspace{-0.2cm}
\end{table*}

\section{DATA COLLECTION}%
\label{sec:dataset}
The dataset captures short-term environmental changes in natural and semi-urban settings, with a focus on their impact on localization and relocalization.
More broadly, it supports research in teach-and-repeat navigation, and robustness to real-world variability, providing a foundation for evaluating perception and mapping systems under evolving outdoor conditions.
Data collection began on February 10, 2025, and is performed weekly on Mondays around 9 AM, ensuring consistent intervals and minimizing temporal variability.
This high-frequency revisit enables the study of localized and progressive changes, such as vegetation movement, object displacement, and subtle terrain alterations, within a controlled and repeatable spatial context.
Further information and download access to the dataset can be found at this address \footnote{\href{https://github.com/RaoufDannaoui1/Natural_and_Semi-Urban_Dataset}{github.com/RaoufDannaoui1/Natural\_and\_Semi-Urban\_Dataset}}.

\subsection{Acquisition Platform and Sensors}
\begin{figure*}[ht!]
  \centering
  \begin{subfigure}[b]{0.2\textwidth}
    \centering
    \includegraphics[width=.89 \linewidth]{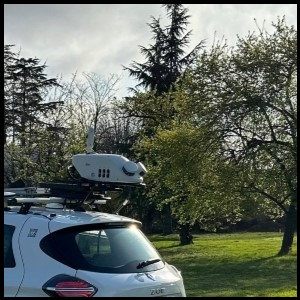}
    \caption{Data acquisition setup}%
	    \label{fig:zoe-leica}
  \end{subfigure}
  \hfill
  \begin{subfigure}[b]{0.38\textwidth}
    \centering
    \includegraphics[width=\linewidth]{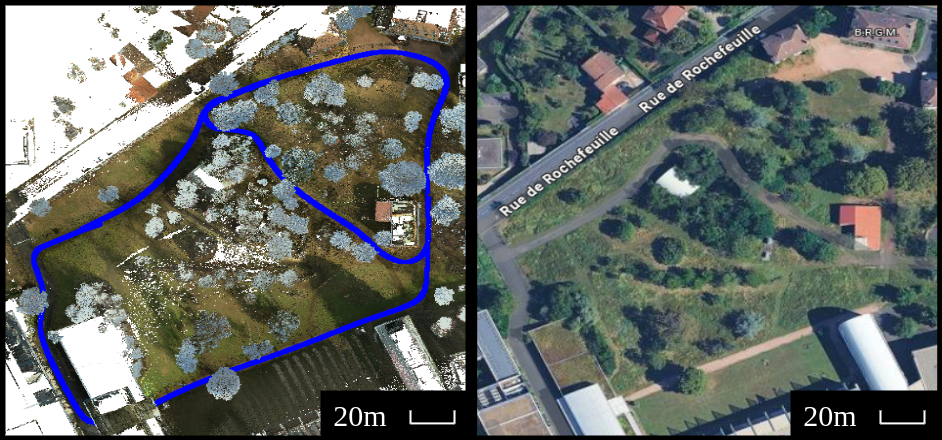}
    \caption{Natural environment}%
    \label{fig:natural-track}
  \end{subfigure}
  \hfill
  \begin{subfigure}[b]{0.38\textwidth}
    \centering
    \includegraphics[width=\linewidth]{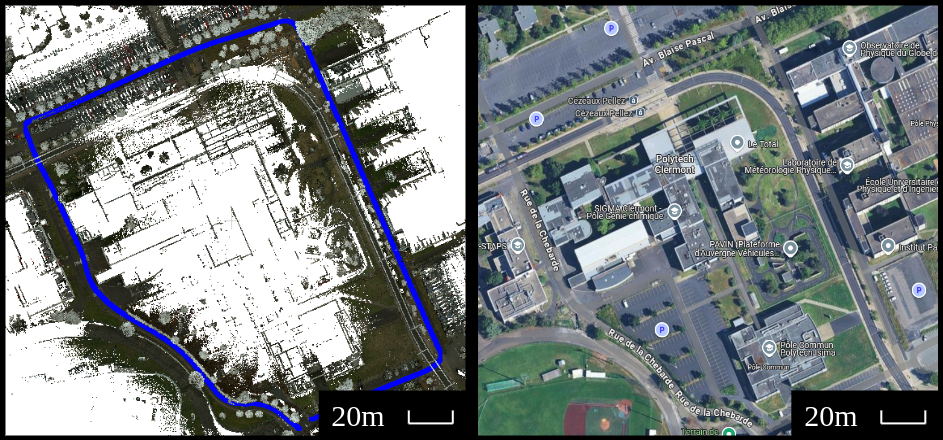}
    \caption{Semi-Urban environment}%
    \label{fig:semi-urban-track}
  \end{subfigure}
\caption{Overview of the acquisition setup and dataset tracks. From left to right: (a) the data acquisition platform consisting of the Leica Pegasus TRK100 mounted on a Zoe electric car, (b) the natural environment track in the INRAE Clermont-Ferrand forest, visualized as a lidar point cloud with trajectory and satellite view of the environment, and (c) the semi-urban track around Universit\'e Clermont Auvergne campus, also shown as a point cloud with trajectory and satellite view of the environment. These two trajectories capture both evolving natural scenes and more stable urban environments, providing a rich basis for evaluating localization performance under varying real-world conditions.}%
  \vspace{-0.4cm}
  \label{fig:dataset_tracks}
\end{figure*}

The data acquisition process is done using a Leica Pegasus TRK100 mobile mapping system, securely mounted on a Zoe electric car as shown in \autoref{fig:zoe-leica}. 
All weekly acquisitions were conducted at consistent driving speeds, resulting in point cloud maps with approximately the same number of points, ensuring that observed variations stem from environmental changes rather than acquisition artifacts.
The vehicle itself serves purely as a transport platform to ensure stable and repeatable trajectories along the predefined paths. 
The Leica Pegasus TRK100 is equipped with two 16-beam lidar sensors, a \SI{360}{deg} panoramic camera, and integrated GNSS/IMU for accurate georeferencing. 
Point clouds are colorized and classified using imagery from the panoramic camera through Leica's post-processing software, which also performs scan alignment using graph-based SLAM to produce aligned maps, as well as automatic anonymization of people and vehicles in the images for privacy reasons.
While color information is preserved in the dataset, it is not used in the current analysis.
Note that the top of the trees are often seen as a light blue because of re-projection errors between the camera and the lidar, therefore associated the color of the sky with the leafs.

\subsection{Environments}
The dataset consists of two distinct trajectories, each selected to capture different types of outdoor environments.
The first path, shown in \autoref{fig:natural-track}, is located in the INRAE Clermont-Ferrand natural park.
It spans approximately \SI{680}{\metre} and is recorded bidirectionally, meaning that the same trajectory is recorded in both directions.
This natural route includes wide range of vegetation, including dense canopy, underbrush, and open clearings. 
Even over short timescales, this forested environment shows noticeable changes such as foliage growth, branch trimming, and ground cover variation, making it well-suited for studying the impact of natural dynamics on relocalization.
The second path, shown in \autoref{fig:semi-urban-track}, runs through the  Universit\'e Clermont Auvergne campus and covers roughly \SI{850}{\metre}, also recorded bidirectionally.
This semi-urban environment combines static man-made features such as buildings, fences, and roads with patches of vegetation and open green spaces.
Bidirectional recording on both tracks ensures a more complete and realistic evaluation setting, capturing the effect of viewpoint variation.

The ground truth map used for localization analysis is extracted from the \texttt{Week 01} dataset, not \texttt{Week 00}. 
This decision was made because the natural environment at INRAE underwent manual trimming between these two sessions. 
Specifically, several trees in the area had dense, overhanging branches in \texttt{Week 00}, which were pruned by \texttt{Week 01}. 
Additionally, \texttt{Week 00} included transient elements such as a parked car that was no longer present in \texttt{Week 01}. 
As our focus is to analyze how environmental changes such as vegetation growth, object displacement, or structural modifications, impact localization over time, it was essential to begin with a clean and stable reference map. 
\texttt{Week 01} offers this stable baseline, free from dense foliage and scene clutter that would otherwise introduce occlusions and inconsistencies unrelated to natural environmental evolution.

The analysis in \autoref{tab:change_percentages} quantifies environmental variation relative to the reference maps: \texttt{Week 01} for the natural track and \texttt{Week 00} for the semi-urban one. 
For each session, we align the point clouds and identify changes as points that are not present in the reference and lie farther than \SI{0.3}{\metre}, are considered changes. 
Around each robot pose, a \SI{35}{\metre} radius sphere is used to compute the percentage of change as the ratio of new points to reference points.

The results reveal distinct behaviors between the two environments.
In the natural track, the detected changes are largely driven by vegetation growth, trimming, or occlusion and are relatively high.
For instance,  in \texttt{Week 00}, several trees had dense, overhanging branches, and a car was present in the scene but not in \texttt{Week 01}, resulting in a median change of \SI{1.81}{\%} and a maximum of \SI{8.68}{\%}, with \texttt{Week 01} used as the baseline (hence excluded from the table).
This level of variation reflects the unstructured and evolving nature of natural environments, where even subtle seasonal or daily shifts can significantly alter the observed scene. 
After the consequent change from \texttt{Week 01} to \texttt{Week 00}, the following weeks showed a lower median, that increase thought weeks.
Notably, \texttt{Week 09} displays a sharp rise in the median change, attributed to a combination of heavy rainfall and warm temperatures during that period, which accelerated vegetation growth and led to visible structural differences in the point clouds.

In contrast, the semi-urban track demonstrates greater consistency across sessions.
\texttt{Week 00} is treated as the reference, and thus no change values are reported for that session.
Following weeks exhibit median changes starting at \SI{2.27}{\%}, with generally moderate maximum values.
This reduced variability is expected in environments dominated by static man-made structures such as buildings and fences.
Nonetheless, transient elements such as parked vehicles or localized vegetational changes still introduce measurable differences as we can see in \texttt{Week 07} where max change was \SI{9.53}{\%}, a large truck parked along the pathway, altering the scene structure and increasing the detection of changes.

These findings underscore the differing challenges for relocalization: the natural environment presents a moving target due to its organic variability, whereas the semi-urban setting offers predictability but still contains patches of vegetation that introduce subtle, localized changes. 
Furthermore, the semi-urban environment also propose numerous unpredictable changes such as the parked car alongside the road. 

\begin{table}[ht!]
  \centering
  \caption{Change percentage summary for natural and semi-urban runs.}
  \begin{tabularx}{1\linewidth}{@{}l 
      >{\centering\arraybackslash}X 
      >{\centering\arraybackslash}X 
      >{\centering\arraybackslash}X 
      >{\centering\arraybackslash}X 
      >{\centering\arraybackslash}X 
      >{\centering\arraybackslash}X @{}}
      \toprule
      \multirow{2}{*}{\textbf{Week}} & 
      \multicolumn{3}{c}{\textbf{Natural [\%]}} & 
      \multicolumn{3}{c}{\textbf{Semi-Urban [\%]}} \\
      \cmidrule(lr){2-4} \cmidrule(lr){5-7}
      & \textbf{Median} & \textbf{IQR} & \textbf{Max} 
      & \textbf{Median} & \textbf{IQR} & \textbf{Max} \\
      \midrule
      Week 00 & 1.81 & 2.64 & 8.68 &  --  &  --  &  -- \\
      Week 01 &  --  &  --  &  --  & 2.27 & 2.65 & 5.62 \\
      Week 02 & 0.32 & 0.15 & 0.68 & 1.73 & 2.10 & 6.19 \\
      Week 03 & 0.81 & 0.34 & 2.62 & 2.20 & 2.39 & 5.28 \\
      Week 04 & 0.65 & 0.57 & 2.56 & 2.30 & 2.09 & 7.92 \\
      Week 05 & 0.81 & 0.57 & 1.49 & 2.78 & 2.44 & 5.64 \\
      Week 06 & 0.89 & 0.58 & 3.31 & 2.35 & 1.49 & 6.71 \\
      Week 07 & 0.46 & 0.24 & 1.30 & 2.55 & 4.32 & 9.53 \\
      Week 08 & 0.72 & 0.38 & 1.29 & 3.09 & 1.80 & 6.46 \\
      Week 09 & 1.63 & 1.88 & 4.03 & 2.42 & 3.61 & 8.56 \\
      \bottomrule
  \end{tabularx}
  \label{tab:change_percentages}
\end{table}

\begin{figure*}[ht]
  \centering
  \includegraphics[width=\linewidth]{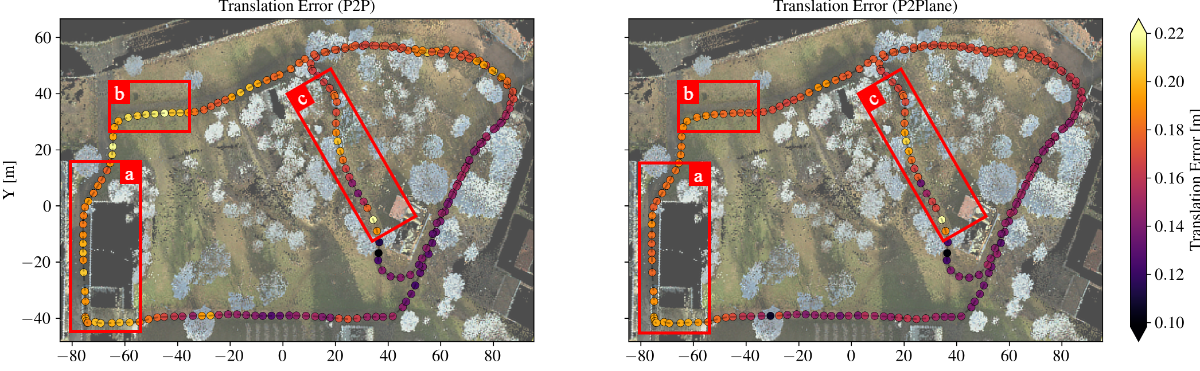}
  \includegraphics[width=\linewidth]{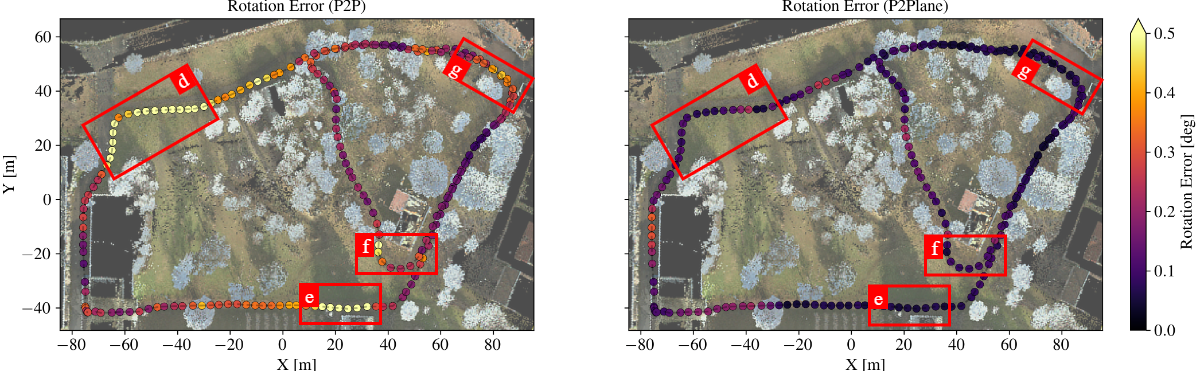}
  \caption{Visualization of trajectory error magnitudes, with annotated key points highlighting translation and rotation error cases. Each labeled region corresponds to a specific environmental challenge affecting each of \ac{ICP} variants performance.}%
  \vspace{-0.4cm}
  \label{fig:trans_rot_error}
\end{figure*}

\section{LOCALIZATION ANALYSIS}%
\label{sec:localization_analysis}
To evaluate how short-term environmental changes impact relocalization performance, we leverage the proposed dataset and evaluate the impact of the changes on the \ac{ICP} algorithm.
The setup focuses on the natural track of the dataset.

\subsection{Methodology}
For each pose, we extract a 35-meter radius spherical submap centered on the robot's pose. 
This ground truth point cloud map is built from the high-density, Leica-aligned 3D data and lightly downsampled using a voxel size of \SI{0.1}{\metre}. 
Subsequent sessions, such as \texttt{Week 00} (before trimming) and \texttt{Week 01} (early regrowth), are used for evaluating short-term changes and their effect on relocalization accuracy.
Projected lidar scans are generated from the aligned point cloud of \texttt{Week-$n$}, designed to emulate a realistic 32-beam sensor with a \SI{360}{deg} field of view, a vertical reach of \SI{15}{\metre}, and a maximum sensing radius of \SI{30}{\metre}. 
While the ground truth map includes the full, unobstructed environment, the projected lidar data respects occlusion constraints.
Each projected scan contains approximately \num{15,000} points.

Although both the projected scan and the ground truth submap are inherently well-aligned, we introduce controlled perturbations to model the localization error typical in real-world systems. 
Specifically, each projected scan is independently perturbed, such that the initial guess for the \ac{ICP} algorithm is
\begin{equation}
    \bm{T}_\text{init} = \exp\left(\bm{\xi}^\wedge\right)\bm{T}_\text{gt}, \quad \bm{\xi}\sim\mathcal{N}\left(\bm{0}, \bm\Sigma_{\bm\xi}\right),
    \label{eq:init}
\end{equation}
using the notations from \cite{bib:barfoot2024state}, $\bm{T}_\text{gt}$ being the ground truth pose, and with ${\bm\Sigma_{\bm\xi}=\text{diag}(0.1^2, 0.1^2, 0.1^2, 0.09^2, 0.09^2, 0.09^2)}$, that is a translation uncertainty of \SI{0.1}{\metre} and rotational uncertainty of \SI{5}{\deg}.
Additionally, the reading scan point cloud is perturbed with a gaussian noise of uncertainty $\sigma = \SI{0.1}{\m}$, simulating sensor uncertainty.

\subsection{ICP Configuration and Experimental Protocol}
To realign each noisy projected scan with the ground truth submap, we apply two variants of the Iterative Closest Point (\ac{ICP}) algorithm: Point-to-Point (PtP) and Point-to-Plane (PtPlane). 
Both implementations rely on the \ac{lpm} library~\cite{bib:pomerleau_comparing_2013}. We configure these \ac{ICP} methods with a maximum correspondence distance of \SI{0.7}{\metre} and allow up to 150 iterations, ensuring convergence even in challenging regions. To increase robustness against outliers, a Cauchy outlier filter is applied.
Each scan is registered against the corresponding ground truth submap using 30 randomized \ac{ICP} initializations using \autoref{eq:init}.
For each frame, the final error is reported as the median translation and rotation error.
The translatation error is simply the norm of the difference of translations between the \ac{ICP} result and the ground truth, while the rotational error is computed as
\begin{equation}
    e_\theta = \left\lVert\log\left( \bm{C}_\text{gt}\bm{C}_\text{\ac{ICP}}^{-1} \right)^\vee\right\rVert,
\end{equation}
again using the notations from \cite{bib:barfoot2024state}, and where $\bm{C}_\text{gt}$, $\bm{C}_\text{\ac{ICP}}$ are respectively the rotational components of the ground truth pose $\bm{T}_\text{gt}$ and \ac{ICP} result  $\bm{T}_\text{\ac{ICP}}$.
No additional downsampling, filtering, or pre-processing is performed after the perturbations to preserve the raw conditions for evaluation.

The experiment is conducted on the natural trajectory, covering approximately \SI{680}{\metre} with about $200$ poses. For this initial study, we compare scans from a single \texttt{Week 00} against the reference map at \texttt{Week 01}. 
This baseline provides a controlled scenario for evaluating how \ac{ICP} methods behave in the presence of real short-term environmental changes. 
Future extensions will include comparisons across multiple weeks to assess cumulative change and localization drift over time.

\subsection{Results}%
\label{sec:ICP_results}
This study examines the performance of Point-to-Point (PtP) \ac{ICP} versus Point-to-Plane (PtPlane) \ac{ICP}, with error analysis visualized across different environmental contexts. 

In terms of trajectory-wide error, PtPlane \ac{ICP} consistently achieves stable results, with translation error remaining within the range of \qtyrange{0.1}{0.2}{\metre} and rotational error below \SI{1}{deg}. 
In contrast, PtP \ac{ICP} translation errors occasionally reaching \SI{0.25}{\metre} and rotational errors peaking near \SI{1}{deg}. 
This highlights the increased sensitivity of PtP \ac{ICP} to environmental complexity and initialization uncertainty. 

\begin{figure}[ht!]
  \centering
  \includegraphics[width= \linewidth]{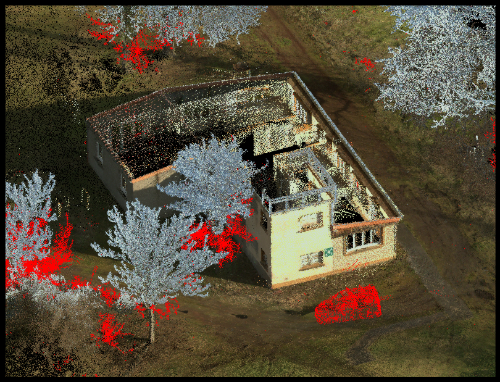}
  \caption{Example of changes in the natural environment between two sessions (\texttt{Week 00} and \texttt{Week 01}). Points in red represent regions where the current scan deviates by more than 0.3 meters from the reference map, indicating newly appeared or displaced objects. In this example, a car has been parked in the area, and several trees have been trimmed between the two acquisitions.}%
  \label{fig:detected_changes_1}
  \vspace{-0.4cm}
\end{figure}

To better understand where and why \ac{ICP} performance degrades, we visualize the entire trajectory colored by median error magnitude, and annotate key points of interest with labeled rectangles, shown in \autoref{fig:trans_rot_error}. 
These areas are further analyzed below, starting with translation error cases then rotational error cases:
\begin{itemize}
    \item Corridor Effect (\textit{a}): The robot navigates between dense vertical vegetation on both sides, creating a corridor-like structure. This leads to ambiguous lateral correspondences and degraded alignment, especially for PtP. A similar limitation is discussed in~\cite{bib:barilKilometerscaleAutonomousNavigation2022}.
    \item Low-Feature Ground (\textit{b}): In this segment, only flat ground is visible, lacking discriminative features. \ac{ICP} fails to estimate displacement accurately due to the absence of meaningful geometry.
    \item Close to Tree Canopy (\textit{c}): The trajectory passes very close to a tree, where partial occlusion and irregular geometry introduce instability in matching. Both methods are affected, but PtPlane performs slightly better.
    \item General Deviation (\textit{d}): In this open-area segment, both methods converge, but PtP yields much higher rotational error compared to PtPlane, indicating reduced robustness to angular drift.
    \item Next to Building (\textit{e} and \textit{g}): The environment offers limited contextual geometry, with only a wall visible, providing limited rotational constraint. While PtPlane remains stable, PtP diverges significantly. 
    \item Sudden Object Change (\textit{f}): A car and new tree branches appears in the scene, depicted in \autoref{fig:detected_changes_1}. PtPlane successfully adapts to this change, while PtP fails to realign effectively, likely due to the object's geometric discontinuity.
\end{itemize}

Overall, these results emphasize the importance of environmental structure in \ac{ICP} performance. 
PtPlane \ac{ICP} demonstrates better generalization in both structured (e.g., walls) and semi-structured (e.g., trees, foliage) areas. 
In contrast, PtP \ac{ICP} struggles under low-feature, ambiguous, or dynamic conditions, reinforcing the need for more robust registration strategies in outdoor relocalization tasks.

\section{ANALYSIS AND DISCUSSION}%
\label{sec:analysis}
\subsection{Error Trends and Patterns}
Our experimental results reveal a consistent advantage of PtPlane \ac{ICP} over PtP \ac{ICP} in both translational and rotational accuracy. 
Across the full trajectory, PtPlane achieved a stable translation error in the range of \qtyrange{0.1}{0.2}{\metre}, while PtP exhibited a wider range, occasionally reaching up to \SI{0.25}{\metre}. 
Similarly, PtPlane maintained rotation errors well under \SI{1}{deg}, while PtP showed greater sensitivity to local geometry, peaking at the upper bound. 
These results highlight the increased robustness of PtPlane \ac{ICP} in semi-structured, evolving environments. 

\subsection{Environmental Impact}
As presented in the annotated trajectory map in Results Section~\ref{sec:ICP_results}, different zones along the path revealed distinct \ac{ICP} behaviors depending on structure and scene dynamics.

In feature-sparse regions, such as open ground with little vertical variation, both PtP and PtPlane struggle due to limited geometry, though PtPlane remains more stable. 
In contrast, structured regions, such as building facades or dense tree clusters, offer stronger constraints, especially for PtPlane due to its use of surface normals.

Consistent with the observations reported in~\cite{bib:pomerleau_comparing_2013}, our experiments confirm that symmetric corridors, formed by vegetation or walls, cause ambiguous correspondences for PtP, leading to drift or misalignment. 
PtPlane, while affected, handles these cases more robustly due to its richer geometric model.

Additionally, dynamic changes also introduce inconsistencies. In such cases, PtPlane tends to rely on stable background geometry, while PtP is more sensitive to new structures.

\subsection{Method Robustness}
Overall, Point-to-Plane \ac{ICP} shows robustness in variety of scene types, making it more suitable for use in natural or semi-structured outdoor environments. 
Its reliance on surface normals provides additional local constraints, improving convergence behavior and reducing the likelihood of local minima, particularly in under-constrained regions.

In contrast, Point-to-Point \ac{ICP} is more vulnerable to initial pose error, occlusion, and structural ambiguity. 
While still effective in scenes with dense and distinct features, its performance quickly degrades in low-salience areas or in the presence of dynamic elements.

\section{CONCLUSION}
We introduced a short-term dataset captured weekly in natural and semi-urban environments to study how real-world changes, such as vegetation trimming and object displacement, affect lidar-based relocalization.
Using projected scans point clouds, we compared two \ac{ICP} variants: Point-to-Point and Point-to-Plane.
Results show that Point-to-Plane consistently outperforms Point-to-Point in both translation and rotation accuracy, especially in areas with repetitive, ambiguous, or sparse geometry.
Trajectory-level error analysis revealed that structural features, feature density, and environmental dynamics are key factors influencing localization, often underrepresented in current benchmarks.
These findings highlight the benefits of incorporating surface normals and support Point-to-Plane \ac{ICP} as a more robust choice for localization in dynamic outdoor settings.

Future work will focus on extending the dataset to cover a full year and all seasons, enabling the study of long-term environmental effects. 
We also plan to evaluate additional registration methods and incorporate scene segmentation to predict localization failures.
Furthermore, the dataset will be used to predict both environmental changes and GNSS coverage variability, particularly by analyzing canopy surface evolution, allowing the robot to anticipate challenging areas where GPS-based localization algorithms are likely to fail.

\section*{ACKNOWLEDGMENT}
This work was supported by the International Research Center “Innovation Transportation and Production Systems” of the I-SITE CAP 20-25. We thank Laurent Malaterre for his assistance in dataset acquisition and Jean Laneurit for his work on the projected lidar scans.

\section*{REFERENCES}
\renewcommand*{\bibfont}{\footnotesize}
\printbibliography[heading=none]

\end{document}